\documentclass[11pt]{article}

\usepackage[final]{acl}

\usepackage{times}
\usepackage{latexsym}

\usepackage[T1]{fontenc}

\usepackage[utf8]{inputenc}

\usepackage{microtype}

\usepackage{inconsolata}

\usepackage{graphicx}
\usepackage{booktabs}
\usepackage{multirow}
\usepackage{amsmath}
\usepackage{algorithm}
\usepackage{algorithmic}
\usepackage[most]{tcolorbox}
\usepackage{float}
\usepackage{caption}
\usepackage{enumitem}
\usepackage{hyperref}

\title{From Confusion to Clarity: Confusion-Aware Retrieval and Knowledge Injection for Text Classification}

\author{
  Manish Gupta\thanks{Equal contribution.} \quad
  Chaitanya Giri\footnotemark[1] \quad
  Jayasimha Talur \\
  Amazon \\
  \texttt{\{manishgp, girichai, talurj\}@amazon.com}
}

\begin{document}
\maketitle

\begin{abstract}
Large language models (LLMs) struggle to classify text into taxonomies with many semantically similar labels, as the distinctions are domain-specific and not captured by pre-training.
To handle large label spaces, a common approach retrieves top-$K$ candidate labels by embedding similarity and prompt the LLM to choose among them.
However, top-$K$ retrieval reduces the number of candidates but does not help the model tell similar ones apart. When two similar labels both appear as candidates, the model lacks the signal to choose correctly between them.
We propose a framework that (1) identifies which label pairs the model struggles to distinguish, (2) expands the candidate set to include confusable labels, and (3) generates targeted rules to differentiate between similar candidates.
The framework requires no fine-tuning, and the generated rules transfer to smaller, cheaper models. On three benchmarks (WOS, Flipkart, LEDGAR), our approach improves Macro F1 by up to 10.0pp over retrieval baselines, with smaller models (2B--20B) gaining up to 11.5pp via cross-model transfer.
\end{abstract}

\section{Introduction}

\begin{figure}[t]
  \centering
  \includegraphics[width=0.9\columnwidth]{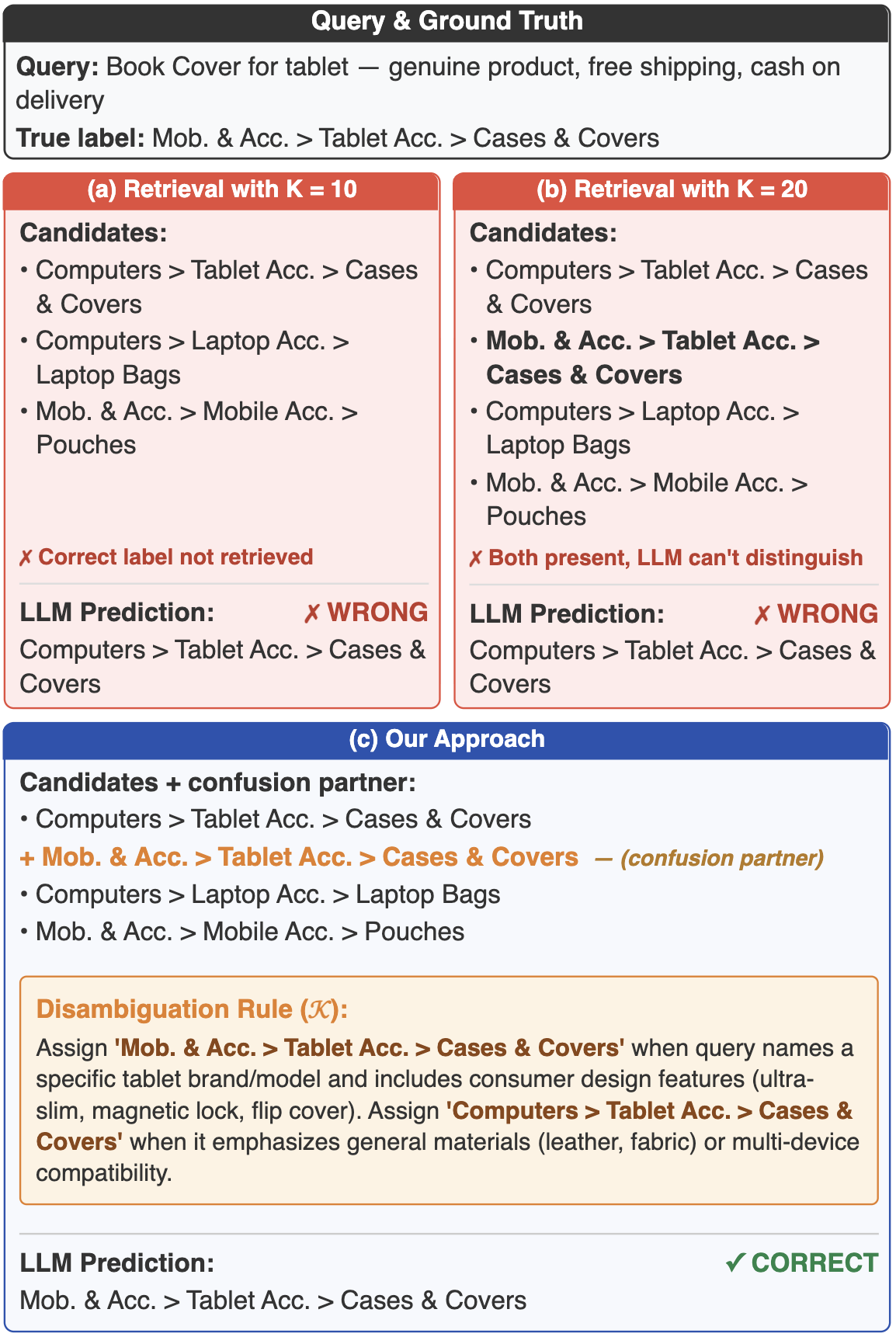}
  \caption{Example illustrating the benefit of confusion-aware knowledge augmentation. Confusion partners recover missing labels, while learned disambiguation rules help the LLM distinguish semantically similar candidate labels}
  \label{fig:framework_approach_fig}
\end{figure}

Classifying text into taxonomies with hundreds of semantically similar labels is a common industrial requirement, yet remains error-prone even for large language models. Product catalogs, customer service queues, and legal document taxonomies often involve labels that overlap in meaning, frequently organized into multi-level hierarchies. Misclassification in these settings directly degrades search relevance, ticket routing, and compliance filtering.

Large language models (LLMs) provide a zero-shot alternative to supervised classifiers: they require no task-specific training data and can interpret label semantics without fine-tuning. A common approach retrieves the top-$K$ candidate labels by embedding similarity and prompts the LLM to select the best match from this reduced set.

However, zero-shot LLMs consistently underperform fine-tuned models on domain-specific taxonomies \citep{vajjala2025textclassification, loukas2023making}. The gap is widest when labels overlap in meaning and only taxonomy-specific conventions separate them: for example, distinguishing ``Women's Western Wear'' from ``Women's Ethnic Wear'' requires knowing how a particular catalog defines each category. Retrieval helps by narrowing the label space, but when two confusable labels both appear among candidates, the LLM still lacks the signal to choose correctly (Figure \ref{fig:framework_approach_fig}). Prior work addresses label ambiguity through tournament-style pairwise comparison \citep{lu2024mitigating} or taxonomy-structure injection \citep{zang2025kghtc}. Neither analyzes which specific pairs the model gets wrong, nor generates targeted rules to fix those errors. By targeting the specific label pairs the model actually fails on, our approach directly addresses the root cause rather than relying on structural or generic signals.

This paper presents \textbf{Confusion-Aware Knowledge Augmented Classification}, a framework that: (1) identifies which label pairs the model struggles to distinguish by analyzing classification errors on a training set; (2) ensures both members of each problematic pair appear in the candidate set at inference time; and (3) generates targeted rules that tell the model how to differentiate between them.\\

We make the following contributions:
\vspace{-2pt}
\begin{itemize}[itemsep=0.1pt,leftmargin=*]
\item A candidate augmentation strategy that uses confusion-matrix analysis to ensure systematically confused labels co-occur in the retrieval set.
\item A three-stage knowledge generation pipeline that produces pairwise disambiguation rules from misclassified training examples.
\item Empirical validation on three datasets and seven models (2B--32B) showing that pairwise disambiguation knowledge consistently improves F1 score, and that knowledge generated once by a large model transfers to smaller, cheaper classifiers without retraining.
\end{itemize}

\begin{figure*}[t]
  \centering
  \includegraphics[width=0.9\textwidth]{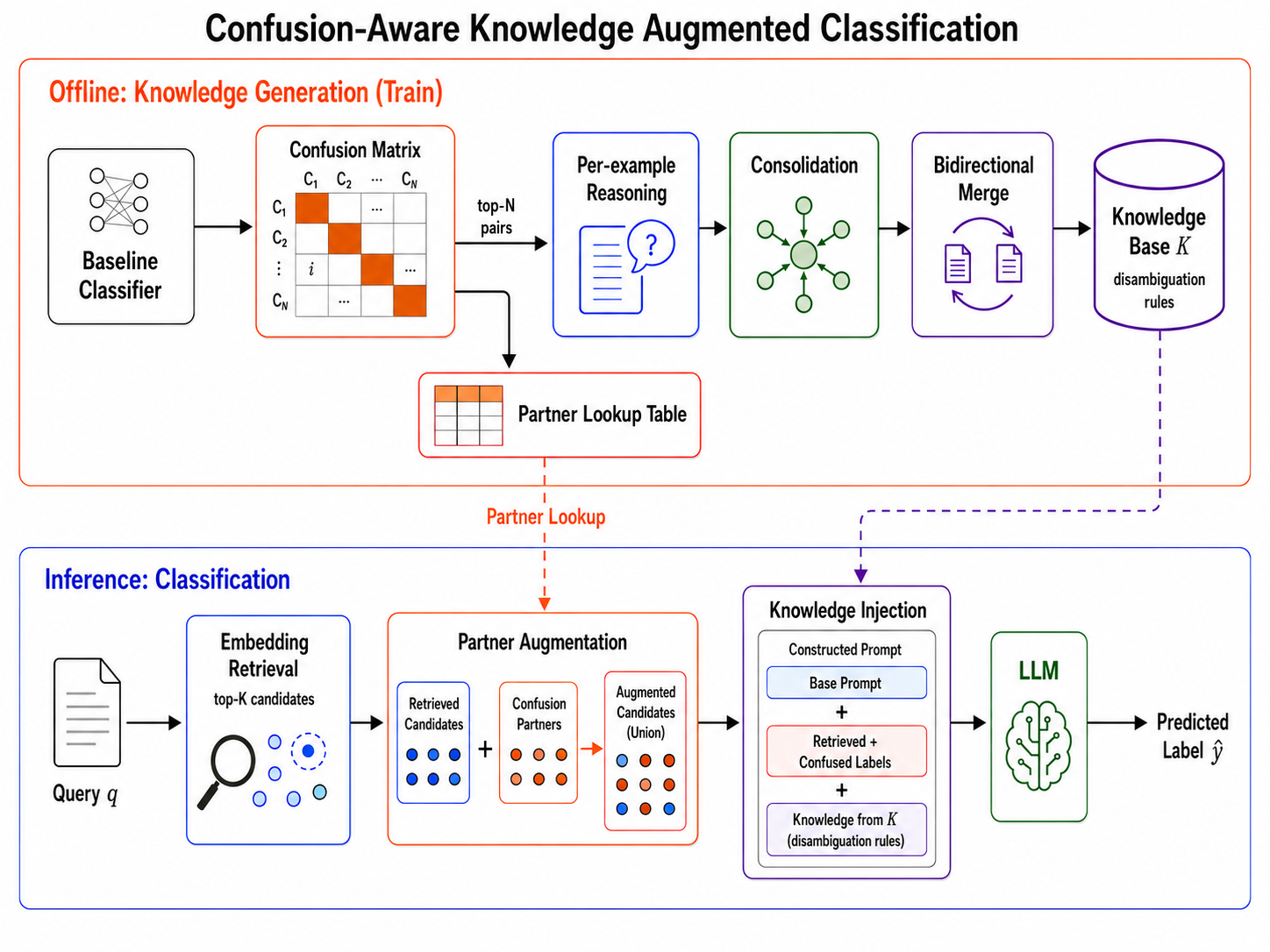}
  \caption{Overview of our framework. The \textit{offline phase} (top) analyzes a baseline classifier's confusion matrix to find systematically confused label pairs and generates pairwise disambiguation rules via a three-stage pipeline. The \textit{inference phase} (bottom) retrieves candidates, augments with confusion partners, injects relevant rules, and predicts.}
  \label{fig:methodology}
\end{figure*}

\section{Related Work}
Our work draws on and extends ideas from LLM-based classification, retrieval augmentation, and prompt optimization.
\paragraph{LLMs for Text Classification.}
Zero-shot LLMs consistently underperform fine-tuned models on domain-specific classification \citep{vajjala2025textclassification, loukas2023making}. However, fine-tuned models require dedicated infrastructure per taxonomy and do not generalize robustly to out-of-distribution inputs. This motivates inference-time approaches: retrieval-based methods narrow the candidate set \citep{pattnaik2025scalable, tabatabaei2025llm, doosterlinck2024incontext}, while others inject structural context such as knowledge graph subgraphs \citep{zang2025kghtc} or LLM-generated taxonomy expansions \citep{paletto2024label}. These methods improve \textit{which} labels reach the LLM; our work addresses what happens \textit{after} retrieval, when similar labels co-occur and the model cannot distinguish them.

\paragraph{Prompt Optimization and Inference-Time Strategies.}
An alternative is optimizing the classification prompt itself. \citet{opsahl2024mipro} optimize instructions and demonstrations for multi-stage LLM programs. \citet{gepa2025} evolve prompts through reflective self-improvement. \citet{lu2024mitigating} reduce label ambiguity through iterative pairwise comparison with hand-crafted descriptions, requiring multiple LLM calls per query. These methods improve prompting generically but do not analyze which specific label pairs cause errors. Our approach is error-informed: we identify problematic pairs from observed misclassifications, generate disambiguation rules specific to those pairs, and apply them in a single LLM call.

\section{Methodology}

Given a text query $q$ and a label taxonomy $\mathcal{L} = \{l_1, \ldots, l_N\}$ with $N$ labels, we formulate the task as assigning $q$ to one label from $\mathcal{L}$. In a zero-shot setting, the LLM receives the query alongside all $N$ labels and their descriptions. As the taxonomy grows, the prompt may exceed the model's context window, and when many labels overlap in meaning, the model struggles to distinguish between them without domain knowledge specific to the taxonomy. A common approach to mitigate the scale problem involves retrieving the top-$K$ labels by embedding similarity, which reduces the candidate space but does not help the model distinguish between similar labels.

Our approach addresses these gaps in three phases (Figure~\ref{fig:methodology}): (1) augmenting the candidate set with labels the model has historically confused with retrieved ones, increasing the chance of correct label being present, (2) generating pairwise disambiguation rules offline from the classifier's own errors, and (3) injecting these rules at inference time to supply the domain knowledge needed to distinguish between similar candidates. We describe each phase in detail below.

\subsection{Confusion-Aware Retrieval}
\label{sec:retrieval}

Standard top-$K$ retrieval uses a general-purpose embedding model that may not reflect taxonomy-specific similarity. Two labels that are close within a taxonomy can be distant in embedding space, causing retrieval to surface one while missing the other. Increasing $K$ recovers missing labels but also admits irrelevant candidates that degrade classification (Figure~\ref{fig:topk_partners}). We exploit a simple observation: if the model has previously misclassified label $l$ as $l'$, this is direct evidence that both labels are plausible for similar queries and should appear together as candidates. We formalize this as \textit{confusion partners}.

\begin{table*}[t]
\centering
\begin{tabular}{@{}lcccccc@{}}
\toprule
 & \multicolumn{1}{c}{\textbf{LEDGAR}} & \multicolumn{2}{c}{\textbf{WOS}} & \multicolumn{3}{c}{\textbf{Flipkart}} \\ \cmidrule(lr){2-2} \cmidrule(lr){3-4} \cmidrule(lr){5-7}
\textbf{Method} &  & L1 & L2 & L1 & L2 & L3 \\
\midrule
Zero-shot (all labels) & 57.1 & 80.5 & 56.6 & 84.4 & 73.3 & 65.5 \\
Few-shot$^\dag$ & 57.8 & 77.6 & 55.0 & -- & -- & -- \\
MIPROv2 \citep{opsahl2024mipro} & 47.2 & 61.4 & 35.9 & 73.8 & 67.0 & 62.4 \\
GEPA \citep{gepa2025} & 52.4 & 68.3 & 32.7 & 61.4 & 61.1 & 58.0 \\
Retrieval ($K{=}20$) & 56.3 & 78.4 & 52.4 & 91.0 & 81.8 & 72.3 \\
\midrule
\textbf{Ours} (Conf.\ Partners + Knowledge) & \textbf{58.9}\textsuperscript{*} & \textbf{81.5}\textsuperscript{*} & \textbf{57.7} & \textbf{92.5} & \textbf{84.4} & \textbf{76.7}\textsuperscript{*} \\
\bottomrule
\end{tabular}
\caption{Macro F1 (\%) on three datasets with Qwen3-32B. Best per column in \textbf{bold}; component ablation in Appendix~\ref{sec:ablation_table}. $^\dag$Exceeds the context window (351 labels + demonstrations). \textsuperscript{*}Our gain over the second-best method in that column is statistically significant ($p<0.05$, paired bootstrap; Appendix~\ref{sec:significance}).}
\label{tab:main_results}
\end{table*}

In order to identify which pair of labels the model confuses the most, we run the retrieval baseline on the training split and record the misclassifications $\hat{Y}_\text{train}$. Each error in which the true label $l$ is predicted as $l'$ increments a directed confusion count $c(l \rightarrow l')$. We rank pairs by this count and select the smallest set $\mathcal{P}$ whose cumulative count covers at least $\tau\%$ of all training errors.

For each label $l$, we define its confusion partner $\textsc{Partner}(l)$ as the label with the highest combined count of $l$ being mistaken for $l'$ and vice versa, since either case indicates that the model cannot distinguish between the two labels (Algorithm~\ref{alg:offline}, lines 1--3).

To increase the chance of the correct label being present, we augment our candidate set $\mathcal{C}$ by adding $\textsc{Partner}(l)$ for each retrieved label $l \in \mathcal{C}$ (Algorithm~\ref{alg:inference}, lines 1--2). This ensures both labels of a confused pair are present, but co-occurrence alone is not enough: the model still lacks the domain-specific signal needed to tell them apart.

\subsection{Knowledge Generation}
\label{sec:knowledge_gen}

We generate disambiguation rules for each pair in $\mathcal{P}$ through a three-stage offline pipeline (Algorithm~\ref{alg:offline}, lines 4--12; prompts in Appendix~\ref{sec:knowledge_prompts}).

\paragraph{Stage 1: Per-example reasoning.} For every training query where $l$ was misclassified as $l'$, we prompt an LLM: \textit{``What signal in this query indicates it belongs to $l$ rather than $l'$?''} This produces a set of per-example observations $O_{l \rightarrow l'}$.

\paragraph{Stage 2: Consolidation.} Individual observations may be query-specific and noisy. We prompt an LLM to distill $O_{l \rightarrow l'}$ into a single directional rule $r_{l \rightarrow l'}$ that captures the consensus signals (examples in Appendix~\ref{sec:generated_knowledge}).

\paragraph{Stage 3: Bidirectional merge.} Pairs confused in both directions yield two directional rules. We prompt an LLM to merge them into one symmetric rule. The result is a knowledge base $\mathcal{K}$ containing one disambiguation rule per pair in $\mathcal{P}$.

\subsection{Inference}
\label{sec:inference}

Inference combines both components into a single LLM call: confusion partners ensure the right labels are present, and the generated rules tell the model how to choose among them. Given a query, we retrieve candidates $\mathcal{C}$ and expand them with confusion partners into $\mathcal{C}'$ (Section~\ref{sec:retrieval}). We then gather applicable rules $\mathcal{R} \subseteq \mathcal{K}$, inject them into the prompt alongside the candidate labels (Appendix~\ref{sec:knowledge_examples}, Figure~\ref{fig:prompt_pairwise}), and the LLM selects the final label $\hat{y}$ from $\mathcal{C}'$ (Algorithm~\ref{alg:inference}).

\section{Experiments}

We evaluate five questions: (i) Do confusion partners and knowledge injection each improve over retrieval independently? (ii) Do they compose for additive gains across datasets? (iii) Does knowledge from a larger model transfer to smaller classifiers? (iv) Do confusion partners outperform simply increasing $K$? (v) Does pairwise knowledge outperform per-label alternatives?

\subsection{Setup}

\paragraph{Datasets.} We evaluate on three publicly available benchmarks:
  WOS\footnote{\raggedright Web-of-Science (WOS-46985): \url{https://data.mendeley.com/datasets/9rw3vkcfy4/2} distributed under CC BY 4.0 licence.}
  \citep{kowsari2017hdltex},
  Flipkart$^\dag$\footnote{\raggedright Flipkart Products: \url{https://www.kaggle.com/datasets/PromptCloudHQ/flipkart-products} distributed under CC BY 4.0 licence.. $^\dag$Pre-processing details in \ref{sec:dataset_preprocessing}}
  \citep{flipkart2018products}, and
  LEDGAR\footnote{\raggedright LEDGAR: sourced from the LexGLUE benchmark (\url{https://huggingface.co/datasets/coastalcph/lex_glue}); LEDGAR is one of its constituent datasets and distributed under CC BY 4.0 licence..}
  \citep{chalkidis2022lexglue}.

\begin{table}[H]
\centering
\footnotesize
\setlength{\tabcolsep}{4pt}
\begin{tabular}{@{}lccccl@{}}
\toprule
\textbf{Dataset} & \textbf{Labels} & \textbf{Levels} & \textbf{Train} & \textbf{Test} & \textbf{Domain} \\
\midrule
LEDGAR & 100 & 1 & 60k & 10k & Legal \\
WOS & 134 & 2 & 32.8k & 9.5k & Academic \\
Flipkart & 351 & 3 & 12.3k & 3.8k & E-commerce \\
\bottomrule
\end{tabular}
\caption{Dataset statistics. WOS and Flipkart are hierarchical; LEDGAR is flat.}
\label{tab:datasets}
\end{table}

\paragraph{Models.} Our main experiments use Qwen3-32B \citep{yang2025qwen3} as both the knowledge generator and the classifier. To test cross-model transfer, we use the larger Qwen3-235B \citep{yang2025qwen3} as the knowledge generator and apply its knowledge to smaller models: Ministral 3 3B and 8B \citep{mistral2025ministral}, Qwen3.5-2B, 4B, and 9B \citep{qwen2025qwen35}, and GPT-OSS-20B \citep{gptoss2025}. We use Qwen3-Embedding-8B \citep{qwen2025embedding} for retrieval in all experiments.

\paragraph{Hyperparameters.} Retrieval returns $K{=}10$ candidates. Confusion pairs are selected as the fewest directed pairs covering $\geq$75\% of training errors. Appendix~\ref{sec:coverage_sensitivity} reports the coverage sweep, and Appendix~\ref{sec:cost_analysis} reports the resulting generation and inference costs. All LLM calls use temperature 0.\footnote{Qwen3-235B is a mixture-of-experts model whose routing is not fully deterministic at temperature 0; we report standard deviation in Table~\ref{tab:transfer}.}

\section{Results}

Table~\ref{tab:main_results} compares our full approach against five baselines on all three datasets with Qwen3-32B; a component-wise ablation is deferred to Appendix~\ref{sec:ablation_table}. Our method achieves the best Macro F1 in every column, with the largest margins on the deepest, most confusable level (Flipkart L3: +11.2pp over zero-shot and +4.4pp over the strongest baseline, retrieval at $K{=}20$). The gains over the second-best method are statistically significant on LEDGAR, WOS L1, and Flipkart L3 (paired bootstrap, $p<0.05$; Appendix~\ref{sec:significance}).

\paragraph{Baselines.} Zero-shot (all labels in the prompt) is competitive at coarse levels but degrades sharply at fine granularity on the largest taxonomy (65.5 F1 on Flipkart L3). Few-shot edges out zero-shot only on LEDGAR (57.8 vs.\ 57.1); on WOS it is slightly worse at both levels, and it cannot be run on Flipkart, where the input prompt plus 351 labels exceed the context window.
Prompt-optimization baselines operate over the full label set but do not match retrieval-augmented approaches on any column, suggesting that instruction tuning alone does not address the underlying confusion between similar labels.
Our primary evaluation focuses on methods that keep the classifier fixed and adapt through inference-time context. We discuss this deployment choice and compare against fine-tuned models in Section~\ref{sec:finetuning}.

\begin{figure*}[t]
  \centering
  \includegraphics[width=0.8\textwidth]{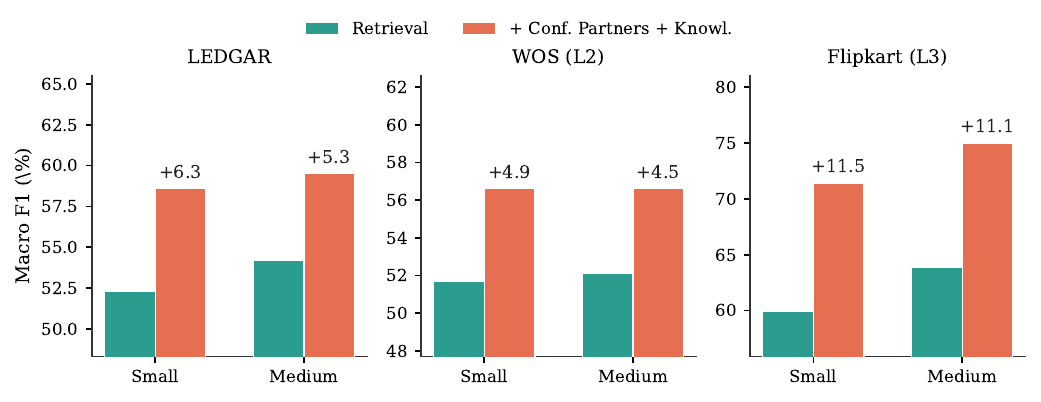}
  \caption{Cross-model transfer: Macro F1 (\%) at each dataset's finest level for retrieval-only vs.\ our full approach, with all knowledge generated by Qwen3-235B. Gains (above each pair) hold across both size groups (Small: 2--4B; Medium: 8--20B) and all datasets.}
  \label{fig:transfer}
\end{figure*}

\paragraph{Error decomposition.} We decompose errors into two failure modes: \textit{retrieval misses} (correct label absent from candidates) and \textit{LLM misses} (correct label present but not selected). On Flipkart L3, retrieval at $K{=}10$ misses the correct label 14\% of the time. Adding confusion partners nearly eliminates this problem (recall: 86\% $\rightarrow$ 98.1\%), but introduces a trade-off: more confusable candidates means more LLM errors (15.5\% $\rightarrow$ 21.9\%). Knowledge injection addresses this, reducing LLM misses to 17.1\%. The two mechanisms are therefore complementary by design. Table \ref{tab:flipkart_retrieval_recall} detail these findings on Flipkart dataset. Refer \ref{sec:error_decomp_appendix} for similar analysis across all datasets.

\begin{table}[h]
\centering
\normalsize
\setlength{\tabcolsep}{5pt}
\begin{tabular}{@{}lcccccc@{}}
\toprule
 & \multicolumn{3}{c}{\textbf{Flipkart}} \\ \cmidrule(lr){2-4}
\textbf{Method} & L1 & L2 & L3 \\
\midrule
Retrieval ($K{=}10$) & 97.8 & 94.2 & 86.0 \\
+ Conf.\ Partners & \textbf{99.9} & \textbf{99.2} & \textbf{98.1} \\
\midrule
Retrieval ($K{=}20$) & 99.3 & 97.4 & 92.4 \\
\bottomrule
\end{tabular}
\caption{Recall (\%, $\uparrow$) on Flipkart dataset (Qwen3-32B): fraction of queries whose correct label is among the candidates. Knowledge injection leaves the candidate set unchanged and so does not affect recall.}
\label{tab:flipkart_retrieval_recall}
\end{table}

\subsection{Analysis}
\label{sec:analysis}

\paragraph{Confusion partners vs.\ larger $K$.}

\begin{figure}[H]
  \centering
  \includegraphics[width=0.9\columnwidth,height=8cm,keepaspectratio]{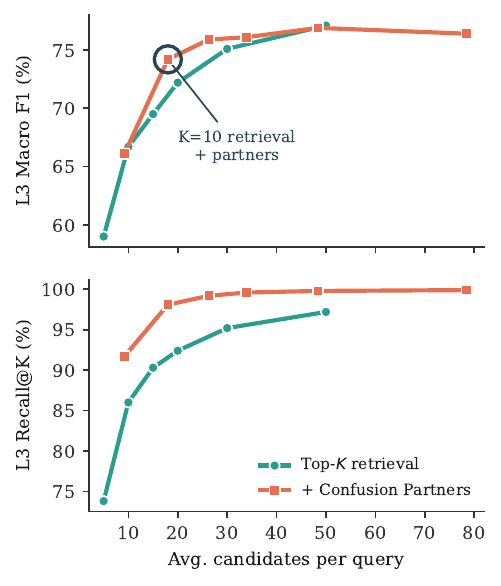}
  \caption{L3 Macro F1 (top) and Recall@K (bottom) on Flipkart (Qwen3-32B) vs.\ average candidates per query. Discussion in Section~\ref{sec:analysis}.}
  \label{fig:topk_partners}
\end{figure}

A natural alternative to confusion partners is simply retrieving more candidates. Figure~\ref{fig:topk_partners} compares both strategies on Flipkart L3, plotting F1 and recall against average candidates per query. Confusion partners at $K{=}10$ (18 candidates on average) achieve 74.2 F1 and 98.1\% recall. Plain retrieval needs $K{=}20$ (20 candidates) to reach only 72.2 F1 and 92.4\% recall, and $K{=}50$ to match the recall. The gap shows that adding labels the model is known to confuse is more effective than retrieving more labels indiscriminately.

\paragraph{Pairwise vs.\ per-label knowledge format.}

Our method presents knowledge as pairwise rules ("Label A vs. Label B: [rule]"). An alternative \textit{per-label} format consolidates all rules for a given label into a single description of its distinguishing features, without naming specific confusers. Table~\ref{tab:knowledge_format} shows pairwise consistently wins. The advantage is that pairwise rules are contextual: the LLM sees both candidates and a rule tailored to exactly that comparison, rather than generic label descriptions that must work against any confuser. This echoes \citet{lu2024mitigating}, who find pairwise comparison more effective than independent label evaluation, though they rely on hand-crafted descriptions (examples of both formats in Appendix~\ref{sec:knowledge_examples}).

\begin{table}[H]
\centering
\setlength{\tabcolsep}{3pt}
\begin{tabular}{@{}lcccccc@{}}
\toprule
 & \multicolumn{2}{c}{\textbf{L1}} & \multicolumn{2}{c}{\textbf{L2}} & \multicolumn{2}{c}{\textbf{L3}} \\ \cmidrule(lr){2-3} \cmidrule(lr){4-5} \cmidrule(lr){6-7}
\textbf{Format} & \textbf{F1} & \textbf{Miss} & \textbf{F1} & \textbf{Miss} & \textbf{F1} & \textbf{Miss} \\
\midrule
No knowledge & 90.8 & 6.2 & 81.9 & 13.3 & 74.2 & 21.9 \\
Per-label & 91.4 & 4.8 & 83.5 & 11.3 & 74.8 & 18.5 \\
Pairwise (ours) & \textbf{92.5} & \textbf{3.8} & \textbf{84.4} & \textbf{10.0} & \textbf{76.7} & \textbf{17.1} \\
\bottomrule
\end{tabular}
\caption{Knowledge format ablation on Flipkart (Qwen3-32B); all rows use retrieval + confusion partners. F1: Macro F1 ($\uparrow$); Miss: LLM miss rate ($\downarrow$). Best per column in \textbf{bold}.}
\label{tab:knowledge_format}
\end{table}

\subsection{Cross-Model Knowledge Transfer}
\label{sec:transfer}

Small models are preferred for inference due to lower latency and cost, but they lack the capacity to generate high-quality disambiguation rules. We therefore generate rules offline with a large model and serve a small, fast model at inference time. We run each smaller model on the training set to identify its own confusion pairs; the large model then writes disambiguation rules for those pairs. We test this by generating all knowledge with Qwen3-235B and applying it to smaller models: Small (2--4B) and Medium (8--20B).

Table~\ref{tab:transfer} confirms effective transfer. The pattern from Table~\ref{tab:main_results} reproduces: partners improve recall, knowledge reduces selection errors, and the combination yields the largest gains. Small models improve by +11.5pp on Flipkart L3 over retrieval-only; Medium models by +11.1pp. Gains hold across all datasets including the flat taxonomy (LEDGAR: +6.3pp Small, +5.3pp Medium). These results indicate that the rules capture domain knowledge that transfers across model sizes, decoupling rule quality from inference cost. Appendix~\ref{sec:frozen_kb} tests the stricter setting in which the same confusion pairs and rules are reused unchanged across classifiers.

\subsection{Comparison with Fine-Tuning}
\label{sec:finetuning}

A natural alternative to our inference-time approach is to use the same labeled data for fine-tuning. This process encodes taxonomy-specific information in model parameters, producing a separate model artifact that must be trained, deployed, and maintained for each taxonomy. Changes to the label space or input distribution require this lifecycle to be repeated. Our method instead targets teams that access foundational models through managed APIs rather than operate their own training and serving infrastructure. This deployment setting motivates our comparison with MIPROv2 and GEPA, which optimize prompts without updating model weights. To complement this evaluation, we compare our method with fine-tuning under clean data and three conditions commonly encountered in production.

Specifically, we fine-tune ModernBERT-base~\citep{warner2024modernbert} and RoBERTa-base~\citep{liu2019roberta} on Flipkart L3 and evaluate them alongside our framework under four settings: clean data, test-time perturbations, label noise, and data scarcity. The clean setting uses the complete training and test splits. In each robustness setting, we vary only the condition under study. Table~\ref{tab:finetuning_robustness} summarizes one representative setting for each condition. Appendix~\ref{sec:finetuning_appendix} reports additional results and implementation details.

\begin{table}[t]
\centering
\scriptsize
\setlength{\tabcolsep}{3pt}
\begin{tabular}{@{}lccc@{}}
\toprule
\textbf{Setting} & \textbf{ModernBERT} & \textbf{RoBERTa} & \textbf{Ours} \\
\midrule
Clean data (complete) & 80.0 & \textbf{81.7} & 76.7 \\
Label noise (20\%) & 60.0 & 61.0 & \textbf{72.1} \\
Data scarcity (10\%, $\sim$3/class) & 23.2 & 31.8 & \textbf{70.2} \\
\midrule
Test-time perturbations (mean $\Delta$) & $-17.0$ & $-9.0$ & $\mathbf{-5.0}$ \\
\bottomrule
\end{tabular}
\caption{Flipkart L3 Macro F1. Rows above the midrule are absolute scores. The final row is the mean change ($\Delta$) across nine test-time perturbations, where values closer to zero are better.}
\label{tab:finetuning_robustness}
\end{table}

On clean data, ModernBERT and RoBERTa reach 80.0 and 81.7 Macro F1, compared with 76.7 for our framework. This advantage, however, does not persist across the robustness settings. With 20\% of the training labels replaced, our framework retains 72.1 Macro F1, compared with 60.0 for ModernBERT and 61.0 for RoBERTa. With only 10\% of the training examples, our framework reaches 70.2, compared with 23.2 and 31.8. Over nine perturbed test sets, our framework loses 5.0 points on average, while ModernBERT and RoBERTa lose 17.0 and 9.0 points. Fine-tuning performs best with clean, abundant data and a stable test distribution, but degrades sharply when these assumptions do not hold in production. Our framework retains substantially more accuracy without parameter updates.

\subsection{Human Evaluation of Generated Rules}
\label{sec:human_eval}

LLM-generated rules may contain unsupported or example-specific distinctions. Our pipeline mitigates this risk through Consolidation (Stage 2), which retains signals that recur across labeled examples, and Bidirectional Merge (Stage 3), which reconciles evidence from both confusion directions when available. These stages reduce example-specific noise, but neither they nor downstream classification gains establish the correctness of individual rules. We therefore evaluate rule quality directly through human review.

We randomly sample 50 Flipkart rules generated by Qwen3-32B. For each rule, the authors review the generated rule alongside the two label paths and example queries from both labels. All three authors independently judge whether each rule correctly distinguishes the two labels in both directions and reject it if it is incorrect, misleading, incomplete, or unhelpful. Together, these assessments produce 150 judgments (Appendix~\ref{sec:human_eval_appendix}). On average, the three authors accept 87.3\% of the sampled rules. High-stakes applications should nevertheless require domain-expert approval before generated rules are deployed.

\section{Conclusion}

We presented a framework that turns LLM classification errors into targeted disambiguation rules, improving accuracy without fine-tuning. Two mechanisms work together: confusion partners ensure the right labels are present as candidates, and pairwise rules help the model choose between them. On three benchmarks, the approach improves Macro F1 by up to 10.0pp, and rules generated by a large model transfer to smaller classifiers for gains of up to 11.5pp. The framework requires only a one-time offline error analysis and generalizes across hierarchical and flat taxonomies.

\section*{Limitations}

The framework has two main limitations. First, the offline phase requires labeled training data and can generate rules only for confusion pairs observed in the baseline errors. If the taxonomy or the model's confusion patterns change, the offline phase may need to be rerun for the affected pairs. Furthermore, extending pair coverage captures more confusions but also lengthens the inference prompt and increases cost, and we have not tested how this balance changes for multilingual inputs, substantially larger taxonomies, or temporal drift. Second, generated rules are not automatically verified. The Consolidation and Bidirectional Merge steps reduce noise from individual examples but do not guarantee that the resulting rule is correct. Although 87.3\% of sampled rules met our criteria for correctness and usefulness, applications with serious consequences still require review by domain experts.

\bibliography{custom}

\appendix

\section{Appendix}
\label{sec:appendix}

\begin{table*}[t]
\centering
\begin{tabular}{@{}lcccccc@{}}
\toprule
 & \multicolumn{1}{c}{\textbf{LEDGAR}} & \multicolumn{2}{c}{\textbf{WOS}} & \multicolumn{3}{c}{\textbf{Flipkart}} \\ \cmidrule(lr){2-2} \cmidrule(lr){3-4} \cmidrule(lr){5-7}
\textbf{Method} &  & L1 & L2 & L1 & L2 & L3 \\
\midrule
Zero-shot (all labels) & 57.1 & 80.5 & 56.6 & 84.4 & 73.3 & 65.5 \\
+ Retrieval ($K{=}10$) & 54.3 & 79.1 & 53.4 & 89.6 & 79.3 & 66.7 \\
+ Confusion Partners & 58.3 & 79.5 & 53.6 & 90.8 & 81.9 & 74.2 \\
+ Knowledge & 55.9 & 80.8 & \textbf{57.8} & 90.7 & 81.3 & 68.8 \\
+ Confusion Partners + Knowledge & \textbf{58.9} & \textbf{81.5} & 57.7 & \textbf{92.5} & \textbf{84.4} & \textbf{76.7} \\
\bottomrule
\end{tabular}
\caption{Component ablation: Macro F1 (\%) with Qwen3-32B, adding each component to the zero-shot baseline. \textit{+ Knowledge} applies knowledge without confusion partners; the last row (our full approach) combines both. Best per column in \textbf{bold}.}
\label{tab:ablation}
\end{table*}

\subsection{Algorithms}
\label{sec:algorithms}

Algorithm~\ref{alg:offline} details the offline knowledge-generation pipeline (Section~\ref{sec:knowledge_gen}) and Algorithm~\ref{alg:inference} the augmented inference procedure (Section~\ref{sec:inference}).

\begin{algorithm}[h]
\caption{Knowledge Generation (Offline)}
\label{alg:offline}
\small
\begin{tabular}[t]{@{}l@{ }l@{}}
\textbf{Input:} & Baseline predictions $\hat{Y}_\text{train}$, coverage threshold $\tau$, \\
 & system prompts: $\text{prompt}_\text{reason}$, $\text{prompt}_\text{consol.}$, $\text{prompt}_\text{merge}$ \\
\textbf{Output:} & Knowledge base $\mathcal{K}$, $\textsc{Partner}(\cdot)$
\end{tabular}
\vspace{2pt}
\begin{algorithmic}[1]
\STATE From $\hat{Y}_\text{train}$, count misclassifications $c(l \rightarrow l')$ for all label pairs
\STATE $\mathcal{P} \leftarrow$ fewest pairs by frequency covering $\geq \tau$\% of total errors
\STATE $\forall\, l$: $\textsc{Partner}(l) \leftarrow \arg\max_{l'} \big(c(l \rightarrow l') + c(l' \rightarrow l)\big)$
\FOR{each $(l \rightarrow l') \in \mathcal{P}$}
  \FOR{each misclassified query $q$ from pair $(l \rightarrow l')$}
    \STATE $O_{l \rightarrow l'} \leftarrow O_{l \rightarrow l'} \cup \{\text{LLM}(\text{prompt}_\text{reason},\; q,\; l,\; l')\}$ \hfill \textit{// per-example reasoning}
  \ENDFOR
  \STATE $r_{l \rightarrow l'} \leftarrow \text{LLM}(\text{prompt}_\text{consol.},\; l,\; l',\; O_{l \rightarrow l'})$ \hfill \textit{// consolidate}
\ENDFOR
\FOR{each unordered pair $\{l, l'\}$ in $\mathcal{P}$}
  \IF{both $r_{l \rightarrow l'}$ and $r_{l' \rightarrow l}$ exist}
    \STATE $\mathcal{K}[l,l'] \leftarrow \text{LLM}(\text{prompt}_\text{merge},\; l,\; l',\; r_{l \rightarrow l'},\; r_{l' \rightarrow l})$ \hfill \textit{// merge}
  \ELSE
    \STATE $\mathcal{K}[l,l'] \leftarrow$ the existing $r_{l \rightarrow l'}$ or $r_{l' \rightarrow l}$
  \ENDIF
\ENDFOR
\end{algorithmic}
\end{algorithm}

\begin{algorithm}[h]
\caption{Augmented Inference}
\label{alg:inference}
\small
\begin{tabular}[t]{@{}l@{ }l@{}}
\textbf{Input:} & Query $q$, knowledge base $\mathcal{K}$, $\textsc{Partner}(\cdot)$, $\text{prompt}_\text{classify}$ \\
\textbf{Output:} & Predicted label $\hat{y}$
\end{tabular}
\vspace{2pt}
\begin{algorithmic}[1]
\STATE Retrieve top-$K$ candidates $\mathcal{C}$ by embedding similarity
\STATE $\mathcal{C}' \leftarrow \mathcal{C} \cup \{\textsc{Partner}(l) \mid l \in \mathcal{C}\}$
\STATE $\mathcal{R} \leftarrow \{\mathcal{K}[l,l'] \mid l, l' \in \mathcal{C}',\; \mathcal{K}[l,l'] \text{ exists}\}$ \hfill \textit{// select rules}
\STATE $\hat{y} \leftarrow \text{LLM}(\text{prompt}_\text{classify},\; q,\; \mathcal{C}',\; \mathcal{R})$ \hfill \textit{// classify}
\end{algorithmic}
\end{algorithm}

\subsection{Dataset Preprocessing}
\label{sec:dataset_preprocessing}

The Flipkart dataset is derived from a publicly available crawl of 20{,}000 product listings~\citep{flipkart2018products}. We retain only the first three levels of the category hierarchy, as deeper levels represent brand or product names rather than semantic categories, and the number of unique labels more than doubles beyond level 3. Labels with fewer than three samples are removed to enable stratified splitting. The final dataset contains 17{,}847 samples across 351 labels. WOS and LEDGAR are used as provided by their respective sources without additional preprocessing.

\subsection{Fine-Tuning Baselines}
\label{sec:finetuning_appendix}

We first compare the methods using the complete, unmodified training and test splits. Table~\ref{tab:finetuning_full} reports Macro F1 at the finest level of each dataset after fine-tuning ModernBERT-base and RoBERTa-base. These clean-data results provide the reference scores for the robustness experiments that follow.

\begin{table}[htbp]
\centering
\scriptsize
\setlength{\tabcolsep}{2.5pt}
\begin{tabular}{@{}lccc@{}}
\toprule
\textbf{Method} & \textbf{Flipkart L3} & \textbf{WOS L2} & \textbf{LEDGAR} \\
\midrule
ModernBERT & 80.0 & 72.9 & 75.4 \\
RoBERTa & 81.7 & 80.4 & \textbf{78.1} \\
\midrule
Ours (Qwen3-32B) & 76.7 & 57.7 & 58.9 \\
\bottomrule
\end{tabular}
\caption{Macro F1 (\%) at each dataset's finest level. Fine-tuned models lead on clean, complete data.}
\label{tab:finetuning_full}
\end{table}

\paragraph{Training details.}
The fine-tuned baselines use a linear classifier over the complete taxonomy path with the AdamW optimizer. Training uses early stopping over at most ten epochs, a maximum sequence length of 512, and a batch size of eight. The classification head uses ten times the pretrained model's learning rate.

\paragraph{Robustness experiments.}
All robustness experiments use Flipkart L3. The test-time perturbation experiment keeps the training data fixed and modifies only the test queries. The label-noise and data-scarcity experiments modify the training data while leaving the test set unchanged. Within each condition, every method uses the same perturbed queries, corrupted labels, or stratified training subset.

\paragraph{Test-time perturbations.}
We use \texttt{nlpaug} 1.1.11~\citep{ma2019nlpaug} to introduce spelling errors at a 10\% rate, word swaps and contextual insertion at 20\%, and word deletion and contextual substitution at 15\%. Keyboard augmentation changes characters in 5\% of words with a 10\% within-word rate. The combined condition applies light spelling, keyboard, and deletion noise. Query shortening keeps four to eight leading product-title tokens, and back-translation sends each query through English--German--English using Amazon Translate.\footnote{\url{https://aws.amazon.com/translate/}} Table~\ref{tab:input_shift_full} reports the change from each method's clean score for all nine conditions.

\begin{table}[htbp]
\centering
\scriptsize
\setlength{\tabcolsep}{2.5pt}
\begin{tabular}{@{}lccc@{}}
\toprule
\textbf{Perturbation} & \textbf{ModernBERT} & \textbf{RoBERTa} & \textbf{Ours} \\
\midrule
Keyboard & $-5.0$ & $-2.0$ & $\mathbf{-1.0}$ \\
Spelling & $-3.0$ & $\mathbf{-2.0}$ & $\mathbf{-2.0}$ \\
Word swap & $-4.0$ & $\mathbf{-1.0}$ & $-3.0$ \\
Word deletion & $-8.0$ & $\mathbf{-4.0}$ & $\mathbf{-4.0}$ \\
Contextual insertion & $-19.0$ & $-5.0$ & $\mathbf{-4.0}$ \\
Contextual substitution & $-29.0$ & $-11.0$ & $\mathbf{-7.0}$ \\
Back-translation & $-22.0$ & $-11.0$ & $\mathbf{-2.0}$ \\
Combined noise & $-9.5$ & $-5.0$ & $\mathbf{-3.0}$ \\
Query shortening & $-57.0$ & $-43.0$ & $\mathbf{-18.0}$ \\
\midrule
\textbf{Mean} & $-17.0$ & $-9.0$ & $\mathbf{-5.0}$ \\
\bottomrule
\end{tabular}
\caption{Robustness to test-time perturbations on Flipkart L3. Each score is the change in Macro F1 from that method's clean result, in percentage points. Values closer to zero are better.}
\label{tab:input_shift_full}
\end{table}

\paragraph{Label noise and data scarcity.}
For label noise, we sample the specified fraction of training examples and replace each selected gold label with a different label drawn uniformly from the other 350 classes. The fine-tuned models train on this corrupted split. Our framework applies the same example-to-label mapping when constructing its confusion pairs and rules. We evaluate all methods on the unchanged test set. For data scarcity, we draw stratified subsets of the training data. The fine-tuned models train on each subset, while our framework discovers its confusion pairs and generates its rules from the same subset.

\paragraph{Training set size and test-time perturbations.}
The preceding experiments vary training-set size and test queries separately. We next vary them together by evaluating all nine perturbed test sets at ten training fractions. At each fraction, our framework discovers the confusion pairs and regenerates the rules from the available training examples. Table~\ref{tab:supervision_sweeps} summarizes representative results for label noise, data scarcity, and their interaction with test-time perturbations. In the joint-robustness rows, each score is the mean Macro F1 across the nine perturbations. The resulting 270 evaluations show how additional labeled data affects robustness rather than performance only on the unchanged test set.

\begin{table}[htbp]
\centering
\scriptsize
\setlength{\tabcolsep}{2.5pt}
\begin{tabular}{@{}llccc@{}}
\toprule
\textbf{Experiment} & \textbf{Setting} & \textbf{ModernBERT} & \textbf{RoBERTa} & \textbf{Ours} \\
\midrule
Label noise & 10\% corrupted & 68.4 & 67.4 & \textbf{73.4} \\
 & 20\% corrupted & 60.0 & 61.0 & \textbf{72.1} \\
 & 50\% corrupted & 41.8 & 40.9 & \textbf{70.1} \\
\midrule
Training data & 10\% ($\sim$3/class) & 23.2 & 31.8 & \textbf{70.2} \\
 & 25\% ($\sim$9/class) & 44.0 & 47.7 & \textbf{69.9} \\
 & 50\% ($\sim$18/class) & 69.8 & \textbf{70.8} & 69.8 \\
\midrule
Joint robustness & 10\% training data & 18.0 & 29.2 & \textbf{67.7} \\
 & 20\% training data & 32.2 & 42.7 & \textbf{68.5} \\
 & 50\% training data & 51.5 & 63.2 & \textbf{69.8} \\
\bottomrule
\end{tabular}
\caption{Representative Flipkart L3 Macro F1 (\%). Label noise and training data use the unchanged test set. Joint robustness reports the mean across nine perturbed test sets.}
\label{tab:supervision_sweeps}
\end{table}

\subsection{Coverage Sensitivity}
\label{sec:coverage_sensitivity}

The coverage threshold controls how much of the observed training error is represented in the knowledge base. We rank directed confusion pairs by frequency and select the smallest set whose cumulative count reaches the threshold. Lower values retain only the most frequent confusions, while higher values admit rarer pairs and make more rules available at inference time.

We sweep the threshold from 10\% to 90\% on each dataset and observe a monotonic increase in Macro F1. Quality is highest at 90\%, but broader coverage makes more rules eligible at inference time and increases the prompt size, so each coverage value carries both a quality and a cost. Table~\ref{tab:coverage_sweep} uses 10\% coverage as the reference. This row reports absolute Macro F1 and defines input-token use as $1.0\times$, while subsequent rows report relative Macro F1 gains and token multipliers. The token counts include the instructions, query, candidate labels, and injected rules, but exclude the one-time generation cost reported in Appendix~\ref{sec:cost_analysis}.

\begin{table}[H]
\centering
\scriptsize
\setlength{\tabcolsep}{3pt}
\begin{tabular}{@{}lrrrrrr@{}}
\toprule
\textbf{Coverage} & \multicolumn{2}{c}{\textbf{Flipkart L3}} & \multicolumn{2}{c}{\textbf{WOS L2}} & \multicolumn{2}{c}{\textbf{LEDGAR}} \\
\cmidrule(lr){2-3}\cmidrule(lr){4-5}\cmidrule(l){6-7}
& \textbf{F1} & \textbf{Tokens} & \textbf{F1} & \textbf{Tokens} & \textbf{F1} & \textbf{Tokens} \\
\midrule
10\% & 0.513 & $1.0\times$ & 0.519 & $1.0\times$ & 0.538 & $1.0\times$ \\
25\% & +2.9\% & $1.1\times$ & +1.5\% & $1.6\times$ & +1.1\% & $1.1\times$ \\
50\% & +8.8\% & $1.4\times$ & +7.9\% & $3.2\times$ & +2.4\% & $1.8\times$ \\
75\% & +21.2\% & $2.2\times$ & +10.4\% & $6.0\times$ & +8.0\% & $3.4\times$ \\
90\% & +31.4\% & $3.6\times$ & +12.5\% & $8.8\times$ & +12.5\% & $6.2\times$ \\
\bottomrule
\end{tabular}
\caption{Sensitivity to confusion-pair coverage. At 10\%, F1 is absolute and token use is normalized to $1.0\times$. Later rows show relative F1 gains and token multipliers.}
\label{tab:coverage_sweep}
\end{table}

We therefore use 75\% as a cost-conscious operating point. Applications can move this threshold according to their accuracy and inference budgets.

\subsection{Knowledge Generation and Inference Cost}
\label{sec:cost_analysis}

Our framework incurs costs at two points: one-time offline knowledge generation and online inference, where confusion partners and applicable rules increase the prompt length. Table~\ref{tab:operational_cost} reports the measured token use for offline generation and the measured overhead at inference.

\begin{table}[H]
\centering
\scriptsize
\setlength{\tabcolsep}{3pt}
\begin{tabular}{@{}lrrrr@{}}
\toprule
\multicolumn{5}{@{}l}{\textit{One-time offline knowledge generation}} \\
\textbf{Dataset} & \textbf{Errors} & \textbf{Pairs} & \textbf{Rules} & \textbf{Tokens (in / out)} \\
\midrule
WOS & 11,363 & 750 & 658 & 10.09M / 1.95M \\
Flipkart & 3,564 & 488 & 455 & 2.55M / 0.71M \\
LEDGAR & 19,528 & 750 & 643 & 12.56M / 2.96M \\
\bottomrule
\end{tabular}
\begin{tabular}{@{}lrrrr@{}}
\toprule
\multicolumn{5}{@{}l}{\textit{Online inference, retrieval only $\rightarrow$ full approach}} \\
\textbf{Dataset} & \textbf{Candidates} & \textbf{Rules} & \textbf{Input tokens} & \textbf{Latency (ms)} \\
\midrule
WOS & 10.0 $\rightarrow$ 19.7 & 54.5 & 1,402 $\rightarrow$ 10,144 & 425 $\rightarrow$ 1,238 \\
Flipkart & 10.0 $\rightarrow$ 18.0 & 28.3 & 1,344 $\rightarrow$ 7,139 & 429 $\rightarrow$ 879 \\
LEDGAR & 10.0 $\rightarrow$ 19.5 & 53.2 & 1,189 $\rightarrow$ 9,248 & 379 $\rightarrow$ 1,110 \\
\bottomrule
\end{tabular}
\caption{Measured offline knowledge-generation cost and online inference overhead. Arrows compare retrieval-only inference with the full approach.}
\label{tab:operational_cost}
\end{table}

The offline phase processes 3,564 to 19,528 errors and produces 455 to 658 rules, consuming 3.26M to 15.52M tokens once per taxonomy. Mean rule length ranges from 85.0 to 91.8 words, with 90th percentiles from 104 to 115 words. At inference, the longer prompts increase median latency from 379--429 ms for retrieval only to 879--1,238 ms for the full approach. This is the principal cost of supplying the additional taxonomy-specific context while retaining a single classification call per query.

\subsection{Reusing a Fixed Knowledge Base}
\label{sec:frozen_kb}

The transfer experiment in Section~\ref{sec:transfer} discovers confusion pairs separately for each classifier, then uses Qwen3-235B to generate their rules. To test stricter reuse, we instead identify the pairs and generate the rules once with Qwen3-235B, then apply the resulting knowledge base unchanged to every classifier. Table~\ref{tab:frozen_kb} compares fixed reuse with model-specific pair discovery.

\begin{table}[H]
\centering
\scriptsize
\setlength{\tabcolsep}{3pt}
\begin{tabular}{@{}lccc@{}}
\toprule
\textbf{Classifier} & \textbf{Retrieval} & \shortstack{\textbf{Fixed}\\\textbf{knowledge}} & \shortstack{\textbf{Model-specific}\\\textbf{knowledge}} \\
\midrule
Ministral-3B & 59.0 & 67.1 & \textbf{72.8} \\
Ministral-8B & 63.9 & 71.5 & \textbf{74.4} \\
GPT-OSS-20B & 64.8 & \textbf{75.2} & 73.0 \\
Qwen3-32B & 66.7 & 73.3 & \textbf{74.7} \\
\bottomrule
\end{tabular}
\caption{Flipkart L3 Macro F1 (\%). Fixed knowledge reuses one Qwen3-235B knowledge base unchanged. Model-specific knowledge discovers pairs for each classifier and uses Qwen3-235B to generate their rules.}
\label{tab:frozen_kb}
\end{table}

Fixed reuse improves Macro F1 by 6.6 to 10.3 points over retrieval alone. Model-specific pairs perform better for three classifiers, while fixed reuse performs better for GPT-OSS-20B. Thus, one knowledge base transfers across classifiers, although model-specific error analysis usually adds further gains.

\subsection{Human Evaluation Protocol}
\label{sec:human_eval_appendix}

We sample 50 rules from the Flipkart knowledge base. Each of the three authors independently reviews every rule, producing 150 judgments. For each item, the annotation interface shows the generated rule, the two complete label paths, up to five example queries from each label, and up to three misclassified queries from each direction with their per-example observations.

Each author judges whether the rule correctly and usefully distinguishes the labels in both directions. A rule is rejected if it is incorrect, misleading, incomplete, or unhelpful. Across the three authors, 87.3\% of the rules meet this standard on average. Since this evaluation cannot guarantee the correctness of every rule, high-stakes applications should require domain-expert approval before deployment.

\subsection{Component Ablation}
\label{sec:ablation_table}

Table~\ref{tab:ablation} isolates the contribution of each component by adding them incrementally to the zero-shot baseline, using Qwen3-32B on all three datasets. Retrieval narrows the label space; confusion partners ensure both members of each confused pair co-occur; knowledge injection supplies pairwise disambiguation rules. The full combination (the last row, equal to ``Ours'' in Table~\ref{tab:main_results}) yields the largest gains, and the two mechanisms are complementary: partners raise Recall@K while knowledge reduces selection errors among candidates (Section~\ref{sec:analysis}).

\subsection{Error Decomposition}
\label{sec:error_decomp_appendix}

Section~\ref{sec:analysis} decomposes errors into recall and LLM misses. Table~\ref{tab:retrieval_recall} reports recall and Table~\ref{tab:llm_miss} the LLM miss rate, both across all three datasets (Qwen3-32B).

\begin{table}[H]
\centering
\small
\setlength{\tabcolsep}{4pt}
\begin{tabular}{@{}lcccccc@{}}
\toprule
 & \textbf{LDG} & \multicolumn{2}{c}{\textbf{WOS}} & \multicolumn{3}{c}{\textbf{Flipkart}} \\ \cmidrule(lr){2-2} \cmidrule(lr){3-4} \cmidrule(lr){5-7}
\textbf{Method} & L1 & L1 & L2 & L1 & L2 & L3 \\
\midrule
Retrieval ($K{=}10$) & 81.7 & 95.1 & 83.4 & 97.8 & 94.2 & 86.0 \\
+ Conf.\ Partners & \textbf{92.8} & \textbf{98.5} & \textbf{90.2} & \textbf{99.9} & \textbf{99.2} & \textbf{98.1} \\
\bottomrule
\end{tabular}
\caption{Recall (\%, $\uparrow$) on all datasets (Qwen3-32B): fraction of queries whose correct label is among the candidates. Knowledge injection leaves the candidate set unchanged and so does not affect recall. LDG = LEDGAR.}
\label{tab:retrieval_recall}
\end{table}

The pattern is consistent: confusion partners raise recall at every level but also increase the LLM miss rate, while knowledge injection lowers the LLM miss rate at each retrieval setting.

\begin{table}[H]
\centering
\small
\setlength{\tabcolsep}{2.5pt}
\begin{tabular}{@{}lcccccc@{}}
\toprule
 & \textbf{LDG} & \multicolumn{2}{c}{\textbf{WOS}} & \multicolumn{3}{c}{\textbf{Flipkart}} \\ \cmidrule(lr){2-2} \cmidrule(lr){3-4} \cmidrule(lr){5-7}
\textbf{Method} & L1 & L1 & L2 & L1 & L2 & L3 \\
\midrule
Retrieval ($K{=}10$) & 16.6 & 17.8 & 31.9 & 5.5 & 10.8 & 15.5 \\
+ Conf.\ Partners & 24.9 & 20.9 & 38.8 & 6.2 & 13.3 & 21.9 \\
+ Knowledge & \textbf{14.2} & \textbf{15.7} & \textbf{26.5} & 3.9 & \textbf{8.5} & \textbf{13.1} \\
+ Conf.\ Partners + Knowl. & 22.5 & 18.7 & 34.0 & \textbf{3.8} & 10.0 & 17.1 \\
\bottomrule
\end{tabular}
\caption{LLM miss rate (\%, $\downarrow$) on all datasets (Qwen3-32B): fraction of in-candidate queries the model classifies incorrectly. Best (lowest) per column in \textbf{bold}. LDG = LEDGAR.}
\label{tab:llm_miss}
\end{table}

\subsection{Statistical Significance}
\label{sec:significance}

For each column of Table~\ref{tab:main_results}, we test whether our full approach (confusion partners + knowledge) significantly beats the \emph{second-best} method in that column. We use a paired bootstrap on the test set: for $B{=}10{,}000$ resamples, we draw test instances with replacement (the same resample applied to both systems), recompute the Macro F1 difference on each resample, and report the 95\% percentile confidence interval and a two-sided $p$-value for the null hypothesis of zero difference. Macro F1 is computed exactly as in the main results; point estimates come from an independent inference run and differ from Table~\ref{tab:main_results} by at most $1.8$ points due to the classifier's non-determinism.

\begin{table}[H]
\centering
\small
\setlength{\tabcolsep}{3.5pt}
\begin{tabular}{@{}llcccc@{}}
\toprule
\textbf{Column} & \textbf{2nd best} & \textbf{$\Delta$F1} & \textbf{95\% CI} & \textbf{$p$} \\
\midrule
LEDGAR & Few-shot & $+1.2$ & $[+0.2, +2.2]$ & $0.016$ \\
WOS L1 & Zero-shot & $+1.2$ & $[+0.5, +1.9]$ & $<0.001$ \\
WOS L2 & Zero-shot & $+0.3$ & $[-0.7, +1.2]$ & $0.600$ \\
Flipkart L1 & Retrieval & $+0.8$ & $[-0.5, +2.5]$ & $0.194$ \\
Flipkart L2 & Retrieval & $+0.6$ & $[-1.0, +2.6]$ & $0.360$ \\
Flipkart L3 & Retrieval & $+2.7$ & $[+1.0, +5.5]$ & $0.005$ \\
\bottomrule
\end{tabular}
\caption{Paired bootstrap significance ($B{=}10{,}000$, Qwen3-32B) of our full approach over the second-best method in each column of Table~\ref{tab:main_results}. $\Delta$F1 is the Macro F1 gain; gains with $p<0.05$ are marked \textsuperscript{*} in Table~\ref{tab:main_results}.}
\label{tab:significance}
\end{table}

Table~\ref{tab:significance} reports the results. Our gains are significant on LEDGAR, WOS L1, and Flipkart L3, where the confidence interval excludes zero. On WOS L2 and the two coarse Flipkart levels, our method remains best but the margin over the second-best method is within sampling noise.

\subsection{Knowledge Generation Prompts}
\label{sec:knowledge_prompts}

The knowledge base $\mathcal{K}$ is produced by the three-stage pipeline of Algorithm~\ref{alg:offline}. Figure~\ref{fig:gen_prompts} lists the system and user prompts used at each stage. All three stages are run once per dataset using the generator model (Qwen3-235B in our cross-model transfer experiments, Qwen3-32B otherwise).

\begin{figure*}[h]
\begin{tcolorbox}[colback=blue!3, colframe=blue!40, title={\small\textbf{Stage 1 Prompt: Per-Example Hint}}, fonttitle=\small]
\footnotesize
\textbf{System:} You are an expert at diagnosing text classification errors. You are shown a query that was misclassified, its correct label, and the wrong label it was assigned. In 2--3 sentences, identify: (1) the specific signal in the query --- terminology, methodology, subject matter, or intent --- that marks it as the correct label; (2) why the wrong label was a tempting but incorrect match (what surface similarity caused the confusion); and (3) a generalizable cue that would help distinguish these two labels on future queries. Be precise and factual, and include domain-specific terms that the classifier may not otherwise know. Output only the observation as plain prose --- no markdown, no lists, no preamble. \\[4pt]
\begin{tcolorbox}[colback=yellow!8, colframe=orange!50, boxrule=0.5pt]
\footnotesize\textbf{User:} Query: \textit{\{query\}}. Correct label: \textit{\{ground\_truth\}} (Description: \{gt\_desc\}). Predicted (wrong) label: \textit{\{predicted\}} (Description: \{pred\_desc\}). This query truly belongs to \textit{\{ground\_truth\}} but was misclassified as \textit{\{predicted\}}. In 2--3 sentences cover (1) the signal marking it as the correct label; (2) why the wrong label looked tempting; (3) a generalizable cue.
\end{tcolorbox}
\end{tcolorbox}

\begin{tcolorbox}[colback=blue!3, colframe=blue!40, title={\small\textbf{Stage 2 Prompt: Per-Direction Consolidation}}, fonttitle=\small]
\footnotesize
\textbf{System:} You are an expert at distilling classification error diagnoses into a reliable decision rule. You are given many per-example observations about why queries belonging to label A were mistaken for label B. Identify the signals that recur across observations and generalize --- ignore one-off or example-specific noise. Then write a boundary: a concrete, actionable rule stating when a query belongs to A rather than B, including the domain-specific facts and terminology a classifier needs to tell them apart. Output valid JSON with fields: \texttt{consensus\_signals} (a list of 3--5 of the most reliable, generalizable discriminators) and \texttt{recommended\_boundary} (a thorough if-then decision rule covering those signals). \\[4pt]
\begin{tcolorbox}[colback=yellow!8, colframe=orange!50, boxrule=0.5pt]
\footnotesize\textbf{User:} Label A (correct): \textit{\{A\}}. Label B (predicted incorrectly): \textit{\{B\}}. Here are \{N\} observations from misclassified examples (queries that belong to A but were predicted as B): \{hints\}. Find the consensus: what signals do most hints agree point toward A and not B? Produce a recommended boundary.
\end{tcolorbox}
\end{tcolorbox}

\begin{tcolorbox}[colback=blue!3, colframe=blue!40, title={\small\textbf{Stage 3 Prompt: Bidirectional Merge}}, fonttitle=\small]
\footnotesize
\textbf{System:} You merge two directional classification rules into a single symmetric decision boundary that will be shown to a classifier at inference time. Rule A$\rightarrow$B explains when a query belongs to A (not B). Rule B$\rightarrow$A explains when a query belongs to B (not A). Produce ONE boundary that captures the key insights from BOTH directions. It must: state the single most reliable factor that separates the two labels; name the misleading surface similarity that causes confusion; and give a clear symmetric rule of the form ``Assign `A' when \ldots; assign `B' when \ldots'' using exact label names in single quotes. Preserve specific signals, keywords, and domain facts. Plain text sentences only; no markdown, lists, or headers. Keep it concise (3--5 sentences). \\[4pt]
\begin{tcolorbox}[colback=yellow!8, colframe=orange!50, boxrule=0.5pt]
\footnotesize\textbf{User:} Label A: \textit{\{A\}}. Label B: \textit{\{B\}}. Rule (when A is correct, not B): \{boundary\_A\}. Key signals for A: \{signals\_A\}. Rule (when B is correct, not A): \{boundary\_B\}. Key signals for B: \{signals\_B\}. Merge these into a single symmetric decision boundary for A vs B.
\end{tcolorbox}
\end{tcolorbox}
\caption{The three system/user prompts used in the offline knowledge generation pipeline (Algorithm~\ref{alg:offline}). The blue box is the system prompt; the highlighted inner box is the user message template with \{placeholders\}.}
\label{fig:gen_prompts}
\end{figure*}

\subsection{Generated Knowledge Examples}
\label{sec:generated_knowledge}

Figure~\ref{fig:gen_knowledge} shows three disambiguation boundaries produced by the pipeline for confused pairs in the Flipkart dataset (Qwen3-32B). Each boundary is a symmetric rule that names the misleading surface similarity and gives the deciding signal for each label.

\begin{figure}[H]
\begin{tcolorbox}[colback=blue!3, colframe=blue!40, title={\footnotesize\textbf{Tablet Cases: Computers vs.\ Mobiles \& Accessories}}, fonttitle=\footnotesize, boxsep=2pt, before skip=0pt, after skip=0pt]
\scriptsize
Assign to \textit{Mobiles \& Accessories} when the query names a specific tablet brand/model (e.g., iPad, Samsung Galaxy) and consumer-design features (ultra-slim, magnetic lock, flip cover). Assign to \textit{Computers} when it emphasizes general materials (leather, fabric), multi-device compatibility, or professional features (keyboard integration, docking). The shared term ``tablet case'' is the source of confusion; brand-model specificity is the deciding signal.
\end{tcolorbox}
\begin{tcolorbox}[colback=blue!3, colframe=blue!40, title={\footnotesize\textbf{Glassware: Bar Glasses vs.\ Glasses \& Tumblers}}, fonttitle=\footnotesize, boxsep=2pt, before skip=0pt, after skip=0pt]
\scriptsize
Assign to \textit{Bar Glasses} when the query emphasizes cocktail/mixed-drink use, aesthetic descriptors, or materials like stainless steel and double-walled construction. Assign to \textit{Glasses \& Tumblers} when it highlights everyday table use, capacity, dishwasher safety, or non-alcoholic beverages (juice, water). Both share the word ``tumbler''; the use-case (spirits vs.\ casual) and construction material decide.
\end{tcolorbox}

\begin{tcolorbox}[colback=blue!3, colframe=blue!40, title={\footnotesize\textbf{Gardening: Plants vs.\ Gardening Tools}}, fonttitle=\footnotesize, boxsep=2pt, before skip=0pt, after skip=0pt]
\scriptsize
Assign to \textit{Gardening Tools} when the query refers to a physical accessory (plant container, grow bag, watering system) with transactional language (price, shipping, model). Assign to \textit{Plants} when it focuses on plant varieties, seeds, germination, or cultivation. The word ``plant'' appears in both; whether the item is an \emph{accessory} or a \emph{living plant} is the deciding signal.
\end{tcolorbox}
\caption{Three generated disambiguation boundaries for confused pairs on Flipkart (Qwen3-32B). Each is a symmetric rule injected into the classification prompt when both labels of the pair appear among candidates.}
\label{fig:gen_knowledge}
\end{figure}

\begin{table*}[t]
\centering
\small
\begin{tabular}{@{}llcccccc@{}}
\toprule
 & & \multicolumn{1}{c}{\textbf{LEDGAR}} & \multicolumn{2}{c}{\textbf{WOS}} & \multicolumn{3}{c}{\textbf{Flipkart}} \\ \cmidrule(lr){3-3} \cmidrule(lr){4-5} \cmidrule(lr){6-8}
\textbf{Size} & \textbf{Method} &  & L1 & L2 & L1 & L2 & L3 \\
\midrule
\multirow{5}{*}{Small (2--4B)} & Zero-shot & 48.5 {\scriptsize($\pm$6.9)} & 78.1 {\scriptsize($\pm$2.4)} & 52.5 {\scriptsize($\pm$3.3)} & 75.5 {\scriptsize($\pm$6.8)} & 64.1 {\scriptsize($\pm$7.3)} & 55.9 {\scriptsize($\pm$7.0)} \\
 & + Retrieval & 52.3 {\scriptsize($\pm$2.3)} & 78.9 {\scriptsize($\pm$0.9)} & 51.7 {\scriptsize($\pm$2.0)} & 87.6 {\scriptsize($\pm$1.5)} & 75.4 {\scriptsize($\pm$1.8)} & 59.9 {\scriptsize($\pm$2.2)} \\
 & + Conf.\ Partners & 54.6 {\scriptsize($\pm$3.0)} & 78.9 {\scriptsize($\pm$1.0)} & 51.2 {\scriptsize($\pm$2.0)} & 88.5 {\scriptsize($\pm$2.0)} & 77.6 {\scriptsize($\pm$3.0)} & 67.2 {\scriptsize($\pm$3.8)} \\
 & + Knowledge & 55.9 {\scriptsize($\pm$1.2)} & 80.8 {\scriptsize($\pm$0.7)} & \textbf{56.6} {\scriptsize($\pm$1.0)} & 89.9 {\scriptsize($\pm$1.1)} & 78.8 {\scriptsize($\pm$1.8)} & 64.0 {\scriptsize($\pm$1.8)} \\
 & + Conf.\ Partners + Knowledge & \textbf{58.6} {\scriptsize($\pm$1.5)} & \textbf{81.0} {\scriptsize($\pm$0.8)} & \textbf{56.6} {\scriptsize($\pm$1.3)} & \textbf{91.0} {\scriptsize($\pm$1.7)} & \textbf{81.4} {\scriptsize($\pm$3.0)} & \textbf{71.4} {\scriptsize($\pm$3.9)} \\
\midrule
\multirow{5}{*}{Medium (8--20B)} & Zero-shot & 54.9 {\scriptsize($\pm$4.2)} & 79.4 {\scriptsize($\pm$2.2)} & 55.0 {\scriptsize($\pm$3.4)} & 84.6 {\scriptsize($\pm$3.1)} & 73.4 {\scriptsize($\pm$2.9)} & 65.6 {\scriptsize($\pm$3.1)} \\
 & + Retrieval & 54.2 {\scriptsize($\pm$1.6)} & 78.4 {\scriptsize($\pm$1.6)} & 52.1 {\scriptsize($\pm$3.3)} & 89.7 {\scriptsize($\pm$0.2)} & 78.6 {\scriptsize($\pm$0.5)} & 63.9 {\scriptsize($\pm$0.8)} \\
 & + Conf.\ Partners & 57.1 {\scriptsize($\pm$2.0)} & 78.2 {\scriptsize($\pm$2.0)} & 51.9 {\scriptsize($\pm$4.3)} & 90.8 {\scriptsize($\pm$0.2)} & 81.3 {\scriptsize($\pm$0.3)} & 72.0 {\scriptsize($\pm$0.6)} \\
 & + Knowledge & 57.5 {\scriptsize($\pm$0.3)} & 80.4 {\scriptsize($\pm$0.3)} & 56.2 {\scriptsize($\pm$0.8)} & 90.7 {\scriptsize($\pm$0.5)} & 80.3 {\scriptsize($\pm$1.5)} & 65.8 {\scriptsize($\pm$0.6)} \\
 & + Conf.\ Partners + Knowledge & \textbf{59.5} {\scriptsize($\pm$1.4)} & \textbf{80.9} {\scriptsize($\pm$0.6)} & \textbf{56.6} {\scriptsize($\pm$1.4)} & \textbf{92.0} {\scriptsize($\pm$0.8)} & \textbf{83.7} {\scriptsize($\pm$1.9)} & \textbf{75.0} {\scriptsize($\pm$1.9)} \\
\bottomrule
\end{tabular}
\caption{Cross-model transfer: Macro F1 (\%) for knowledge generated by Qwen3-235B and applied to smaller classifiers, reported as mean (std) over each size group. Small: Ministral-3B, Qwen3.5-2B/4B; Medium: Ministral-8B, Qwen3.5-9B, GPT-OSS-20B.}
\label{tab:transfer}
\end{table*}

\subsection{Error Analysis}
\label{sec:error_analysis}

To understand the limits of the approach, we analyze the 737 residual errors (19\% of test samples) produced by the best configuration from Table~\ref{tab:main_results} (Qwen3-32B with partners + knowledge) on Flipkart at the finest granularity (L3). Table~\ref{tab:error_breakdown} categorizes these errors by source, and Table~\ref{tab:top_errors} lists the most frequent error pairs.

\begin{table}[H]
\centering
\footnotesize
\begin{tabular}{@{}lcc@{}}
\toprule
\textbf{Error Category} & \textbf{Count} & \textbf{\% of Errors} \\
\midrule
Retrieval miss & 73 & 9.9\% \\
Taxonomy inconsistency & 153 & 20.8\% \\
Semantic overlap (top pairs) & 230 & 31.2\% \\
Long-tail / other & 281 & 38.1\% \\
\bottomrule
\end{tabular}
\caption{Error breakdown for the full pipeline on Flipkart L3 (Qwen3-32B, 737 errors out of 3880 test samples).}
\label{tab:error_breakdown}
\end{table}

\begin{table}[H]
\centering
\footnotesize
\setlength{\tabcolsep}{3pt}
\begin{tabular}{@{}llc@{}}
\toprule
\textbf{Ground Truth (L3)} & \textbf{Predicted (L3)} & \textbf{N} \\
\midrule
Car Interior \& Exterior & Car Interior$^\dagger$ & 114 \\
Western Wear & Fusion Wear & 71 \\
Western Wear & Leggings \& Jeggings & 37 \\
Ethnic Wear & Fusion Wear & 32 \\
Women's Casual Shoes & Men's Casual Shoes & 23 \\
\midrule
\multicolumn{2}{@{}l}{\textit{Top 5 pairs total}} & 277 \\
\multicolumn{2}{@{}l}{\textit{\% of all errors}} & 37.6\% \\
\bottomrule
\end{tabular}
\caption{Top 5 remaining error pairs at L3. $\dagger$Labels in different L2 categories describing the same concept (taxonomy overlap). L2 paths omitted for space.}
\label{tab:top_errors}
\end{table}

Three patterns emerge from the remaining errors:

\paragraph{Taxonomy inconsistency (153 errors, 20.8\%).} Labels that describe the same concept appear under different parent categories. The largest case (114 errors) involves \textit{Accessories \& Spare parts $\rightarrow$ Car Interior \& Exterior} vs.\ \textit{Car Accessories $\rightarrow$ Car Interior}. Similarly, \textit{Showpieces} appears under near-identical parent paths with minor naming variations. These cases reflect taxonomy design choices that are difficult for any classifier to resolve from product text alone.

\paragraph{Genuine semantic overlap (230 errors, 31.2\%).} The Western Wear / Fusion Wear / Ethnic Wear cluster accounts for 155 of these. ``Fusion Wear'' by definition blends Western and Ethnic styles, making many items genuinely ambiguous. Our disambiguation rules reduce this confusion on test (from 112 errors with retrieval-only to 71 with the full pipeline for Western$\rightarrow$Fusion alone), but a residual remains for items that are inherently multi-category. Gender-ambiguous footwear (Women's vs.\ Men's Casual Shoes) contributes another 36 errors.

\subsection{Knowledge Format Examples}
\label{sec:knowledge_examples}

We illustrate both knowledge formats using the confused pair \textit{Girls Wear} vs.\ \textit{Infants Wear} from the Flipkart dataset (351 labels, 3-level hierarchy). The pairwise format (Figure~\ref{fig:prompt_pairwise}) presents both labels side-by-side with an explicit decision rule, while the per-label format (Figure~\ref{fig:prompt_perlabel}) describes each label independently without direct contrast.

\begin{figure*}[p]
\begin{tcolorbox}[colback=blue!3, colframe=blue!40, title={\small\textbf{System Prompt: Pairwise Knowledge Format}}, fonttitle=\small]
\small
You are an expert text classifier. Given a query and a list of candidate labels, select the single most appropriate label for the query. \\[4pt]
How to classify: \\
1. Read the query and identify its primary subject, intent, or contribution. \\
2. Evaluate EACH candidate independently on how well it matches that primary subject. \\
3. When several candidates seem plausible, choose the one that captures the query's CENTRAL focus. \\[4pt]
\ldots \\[4pt]
\begin{tcolorbox}[colback=yellow!8, colframe=orange!50, boxrule=0.5pt]
\small\textbf{DISAMBIGUATION KNOWLEDGE:} \\[3pt]
\textit{`Kids' Clothing $|$ Girls Wear'} vs \textit{`Kids' Clothing $|$ Infants Wear':} \\[2pt]
If the query includes terms like `infant', `newborn', or `baby', or references soft, age-appropriate motifs emphasizing comfort for very young children, classify as \textit{Infants Wear}. If the query refers to `girl's', includes age ranges (e.g., `2--4 years'), or describes fashion-oriented items (e.g., `party wear', `printed top'), classify as \textit{Girls Wear}.
\end{tcolorbox}
\vspace{2pt}
Output ONLY valid JSON: \{``task'': ``...'', ``label'': ``exact label from candidates'', ``confidence'': ``high/medium/low''\}
\end{tcolorbox}
\caption{Classification prompt with \textbf{pairwise knowledge} for the confused pair \textit{Girls Wear} vs.\ \textit{Infants Wear} (Flipkart, Qwen3-32B). The highlighted block shows the contrastive disambiguation boundary injected into the system prompt.}
\label{fig:prompt_pairwise}

\vspace{12pt}

\begin{tcolorbox}[colback=blue!3, colframe=blue!40, title={\small\textbf{System Prompt: Per-Label Knowledge Format}}, fonttitle=\small]
\small
You are an expert text classifier. Given a query and a list of candidate labels, select the single most appropriate label for the query. \\[4pt]
How to classify: \\
1. Read the query and identify its primary subject, intent, or contribution. \\
2. Evaluate EACH candidate independently on how well it matches that primary subject. \\
3. When several candidates seem plausible, choose the one that captures the query's CENTRAL focus. \\[4pt]
\ldots \\[4pt]
\begin{tcolorbox}[colback=green!5, colframe=green!40, boxrule=0.5pt]
\small\textbf{LABEL KNOWLEDGE:} \\[3pt]
\textit{`Kids' Clothing $|$ Girls Wear':} Clothing intended for young girls, featuring age-specific indicators such as ``girl's,'' ``kids,'' or ``children.'' Includes casual, party, or special occasion wear with descriptors like ``cotton,'' ``printed top,'' or ``party wear.'' Child-specific sizing (e.g., ``2--4 years'') confirms this category. \\[4pt]
\textit{`Kids' Clothing $|$ Infants Wear':} Queries with explicit infant-specific terms like ``infant,'' ``newborn,'' or ``baby,'' emphasizing comfort, softness, and suitability for very young children. Key signals include infant product types (e.g., bloomers, bodysuits), fabric descriptors like ``soft cotton,'' and brand names like ``Mothercare.''
\end{tcolorbox}
\vspace{2pt}
Output ONLY valid JSON: \{``task'': ``...'', ``label'': ``exact label from candidates'', ``confidence'': ``high/medium/low''\}
\end{tcolorbox}
\captionof{figure}{Same prompt with \textbf{per-label knowledge} for the same pair (Flipkart, Qwen3-32B). Each label described independently without direct contrast.}
\label{fig:prompt_perlabel}
\end{figure*}

\subsection{Cross-Model Transfer Results}
\label{sec:transfer_table}

Table~\ref{tab:transfer} reports the full per-size-group cross-model transfer results summarized in Section~\ref{sec:analysis} and Figure~\ref{fig:transfer}.

\end{document}